\documentclass[letterpaper]{article} % DO NOT CHANGE THIS
\usepackage{aaai24} 

\usepackage{times}  % DO NOT CHANGE THIS
\usepackage{helvet}  % DO NOT CHANGE THIS
\usepackage{courier}  % DO NOT CHANGE THIS
\usepackage[hyphens]{url}  % DO NOT CHANGE THIS
\usepackage{graphicx} % DO NOT CHANGE THIS
\usepackage{natbib}  % DO NOT CHANGE THIS AND DO NOT ADD ANY OPTIONS TO IT
\usepackage{caption} % DO NOT CHANGE THIS AND DO NOT ADD ANY OPTIONS TO IT
\usepackage{algorithm}
\usepackage{algorithmic}

\usepackage{amsmath}
\usepackage{refcount}
\usepackage{longtable}
\usepackage{subcaption}
\usepackage{multirow}
\usepackage{tablefootnote}
\usepackage{booktabs}
\usepackage{xcolor}
\usepackage{makecell}
\usepackage{tabularx}
\usepackage[title]{appendix}

\usepackage{newfloat}
\usepackage{listings}
\DeclareCaptionStyle{ruled}{labelfont=normalfont,labelsep=colon,strut=off} % DO NOT CHANGE THIS
\floatstyle{ruled}
\newfloat{listing}{tb}{lst}{}
\floatname{listing}{Listing}
\title{Examining the Behavior of LLM Architectures Within the Framework of Standardized National Exams in Brazil}
\author{
    Marcelo Sartori Locatelli, Matheus Prado Miranda, Igor Joaquim da Silva Costa, Matheus Torres Prates, Victor Thomé, Mateus Zaparoli Monteiro, Tomas Lacerda, Adriana Pagano, Eduardo Rios Neto, Wagner Meira Jr., Virgilio Almeida\\
}
\affiliations{
    Universidade Federal de Minas Gerais, Belo Horizonte, Brazil\\
    \{locatellimarcelo, matheus.prado, igor.joaquim, matheus.prates, victor-thome, mateuszaparoli, tomas.muniz, meira, virgilio\}@dcc.ufmg.br, 
    apagano@letras.ufmg.br, eduardo@cedeplar.ufmg.br
}

\begin{document}

%%
%% The "title" command has an optional parameter,
%% allowing the author to define a "short title" to be used in page headers.
% OTHER OPTIONS, using 3 just because it is more vague
% 1. Social and economic bias of LLM arcitectures in the Brazilian context.
\newcommand{\marcelo}[1]{\textcolor{red}{  (\textbf{MSL:} #1)}}
\newcommand{\mprado}[1]{\textcolor{cyan}{  (\textbf{M PRADO:} #1)}}
\newcommand{\mzaparoli}[1]{\textcolor{teal}{  (\textbf{M ZAPAROLI:} #1)}}
\newcommand{\mtorres}[1]{\textcolor{purple}{  (\textbf{M TORRES:} #1)}}
\newcommand{\victor}[1]{\textcolor{olive}{ (\textbf{VICTOR:} #1)}}
\newcommand{\igor}[1]{\textcolor{orange}{  (\textbf{IGOR:} #1)}}
\newcommand{\tomas}[1]{\textcolor{green}{  (\textbf{TOMAS:} #1)}}
% 2. Do LLM architectures reproduce social and economic inequalities in Brazilian Portuguese Language?

%%
%% By default, the full list of authors will be used in the page
%% headers. Often, this list is too long, and will overlap
%% other information printed in the page headers. This command allows
%% the author to define a more concise list
%% of authors' names for this purpose.

 \maketitle
%%
%% The abstract is a short summary of the work to be presented in the
%% article.
\begin{abstract}
The Exame Nacional do Ensino M\'edio (ENEM) is a pivotal test for Brazilian students, required for admission to a significant number of universities in Brazil. The test consists of four objective high-school level tests on Math, Humanities, Natural Sciences and Languages, and one writing essay. Students' answers to the test and to the accompanying socioeconomic status questionnaire are made public every year (albeit anonymized) due to transparency policies from the Brazilian Government. In the context of large language models (LLMs), these data lend themselves nicely to comparing different groups of humans with AI, as we can have access to human and machine answer distributions. We leverage these characteristics of the ENEM dataset and compare GPT-3.5 and 4, and MariTalk, a model trained using Portuguese data, to humans, aiming to ascertain how their answers relate to real societal groups and what that may reveal about the model biases. We divide the human groups by using socioeconomic status (SES), and compare their answer distribution with LLMs for each question and for the essay. We find no significant biases when comparing LLM performance to humans on the multiple-choice Brazilian Portuguese tests, as the distance between model and human answers is mostly determined by the human accuracy. A similar conclusion is found by looking at the generated text as, when analyzing the essays, we observe that human and LLM essays differ in a few key factors, one being the choice of words where model essays were easily separable from human ones. The texts also differ syntactically, with LLM generated essays exhibiting, on average, smaller sentences and less thought units, among other differences. These results suggest that, for Brazilian Portuguese in the ENEM context, LLM outputs represent no group of humans, being significantly different from the answers from Brazilian students across all tests.
\end{abstract}

%%
%% This command processes the author and affiliation and title
%% information and builds the first part of the formatted document.

\section{Introduction}

Over the last few years, advancements in NLP and machine learning have culminated in the creation of large language models (LLMs) that excel in a variety of tasks, such as translating, question-answering, summarization, and natural language inference. As these models are getting more and more adopted by the general public, as well as the industry and academia, researchers from many different backgrounds have started evaluating how they fare on a plethora of benchmarks for distinct subjects and tasks~\cite{chang2023survey}. A significant part of this research effort focuses on benchmarks where there is a desirable outcome, or ground-truth that the model is expected to produce, for example the MMLU dataset~\cite{hendrycks2020measuring}, question answering~\cite{liang2022holistic}, and even some more domain specific tasks, such as standardized exams~\cite{giannos2023performance,nunes2023evaluating}.

While these types of evaluation are important, researchers have recently been shifting their attention towards measuring aspects of LLMs that are not directly related to performance. These come in the form of various fairness related concerns, with recent work showing how models may present racial and gender bias~\cite{cheng2023marked}, political bias~\cite{feng2023pretraining}, among others. One way which researchers used to establish which biases a model may show was by comparing model and human responses on a given questionnaire or benchmark~\cite{santurkar2023whose,gurnee2023language}.

Despite the importance of such work and the wide usage of LLM powered tools such as ChatGPT by people of different countries and cultural backgrounds\footnote{\url{https://explodingtopics.com/blog/chatgpt-users}}, very few studies were conducted considering languages other than English. In this work we attempt to bridge that gap on the Brazilian Portuguese language by utilizing the Exame Nacional do Ensino Médio (ENEM) test, which is taken by millions of students seeking to enter a university every year and covers the subjects learned up to high school. This test is made of two parts: 180 objective questions from four subjects (Math, Languages, Natural Science and Humanities) and an essay.

The idea is that by using the abundant ENEM student data, made available publicly every year, it is possible to compare model and human outputs in the context of the standardized test. This would be a way to evaluate which biases LLMs may (or may not) exhibit in the Brazilian context, as the dataset includes not only information about their answers, but also about their income, city, race, gender, among others. Thus, the models, in this context, would be biased toward a certain group if their generated answers and essay were to more closely resemble that subset of humans' behavior, as defined in \cite{atari2023humans}.

We focus on the social and economic bias, which is specially relevant as Brazil has significant income disparities, making it one of the most unequal countries globally. These inequalities are evident between different regions, especially the wealthier southern regions and the less affluent northern ones. Disadvantaged groups such as female-headed households, Afro-Brazilians, and indigenous populations are disproportionately represented among the poor. Workplace inequality in Brazil is further exacerbated by race and gender. A study by the Brazilian Federal Institute of Geography and Statistics (IBGE) revealed that, on average, White workers earn 75.7\% more than Black and Brown individuals, an ethnic category used in Brazilian censuses for mixed-race residents~\cite{IBGE}. Our study explores LLM models to assess if the key indicators of inequality in Brazil are mirrored in the models using the large data collection from ENEM. Since the language models performance is related to their pre-training data, we opt to use both Portuguese pre-trained models (MariTalk) as well as multilingual models (GPT-3.5 turbo, GPT-4) aiming to answer the following questions:

\begin{itemize}
    \item How do the models stack against one another in the context of ENEM?
    \item Which socioeconomic biases do these models show in the context of ENEM (if any)?
    \item When considering the text generated for the essays, which groups would it belong to if written by a human (if any)?
    \item Are there significant differences between the LLM generated and human written essays? If such differences exist, what are the specific aspects of syntactic complexity that are affected, and to what degree?
\end{itemize}

We find that MariTalk and GPT-4 are generally more proficient than GPT-3.5 in the subjects of Languages and Humanities in Brazilian Portuguese, while both GPT models are far superior in natural sciences. In the Math test, all models show subpar results, with lower than 33\% accuracy when considering those questions.  Despite the models appearing to be closer to the higher socioeconomic status groups, we find little evidence of bias in the multiple-choice tests, as the probability of a LLM getting the answer correct plays a much bigger part in the increase in similarity between model and human answers than their actual socioeconomic status.  

Additionally, we find that the models seem to be closest to humans in tests where it performs worse (Natural Sciences and Math) due to less confident predictions, which leads to more well distributed probability among alternatives, better matching human tendencies.

When comparing the essays generated by the LLMs and human written text, we observe that they are not all that similar, as, despite being pre-trained with human data, the models' choice of words is significantly different from that of the exam takers. This difference manifests itself in a very distinct choice of words by the models coupled with a significant difference in the linguistic properties of the text produced by the LLMs, which show much less variance in the number of words for each essay. Additionally, the machine produced texts differ from humans on a variety of syntactic complexity metrics, generating on average longer clauses, but shorter sentences and less phrases per clause, while also using less T-units per sentence. These combine to form model created text that is different in form and in content when contrasted with human ENEM essays.

This research comes in a context where an increasing number of studies position LLMs as if they were human, applying both objective and subjective human tests or questionnaires~\cite{pan2023llms,atari2023humans} in an attempt to show what kind of human these models would be in our society. By exploring the objective multiple choice questions and the subjective ENEM essay, we seek to expand the understanding of how these models may relate to actual human capabilities. Our findings indicate that, for Brazilian Portuguese, LLM capabilities are inconsistent with humans regardless of their origin or status, showing no biases of this kind. Our findings align with previous work which drew similar conclusions in very different situations~\cite{mahowald2024dissociating,pavlick2023symbols}. Our work complements previous research in this vein by directly contrasting LLMs' outputs with plentiful, real-world, human-generated data.

%With the increase in research exploring the possible similarities between large language models (LLMs) and humans, particularly in personality~\cite{pan2023llms} and cognitive capacities~\cite{leng2023llm}, we thought the approach of using ENEM was a great way to determine whether these models can emulate human-like reasoning and articulation, especially when dealing with Brazilian Portuguese. By examining the models' performance on multiple-choice questions and analyzing their ability to generate coherent and contextually relevant responses, as well as their capacity to produce essays, which are naturally more subjective, we seek to understand how well these models can provide nuanced, contextually aware answers and how those compare to the ones humans provide.

\section{Related Work}

 Due to ENEM's multidisciplinary nature, sporting 45 questions for each of four distinct subjects - Humanities, Math, Natural Sciences, and Languages - the test has been used as a benchmark for models in the Brazilian Portuguese language. For example, in \cite{nunes2023evaluating}, Nunes et al. use questions extracted from ENEM as a way to evaluate GPT-3.5 and GPT-4. They found the latter to be an improvement in performance, especially in conjunction with Chain-of-Tought prompting.  More recently, the test has been included in the Portuguese Evaluation Tasks (Poeta) benchmark, used to compare Sabiá~\cite{pires2023sabi}, a model pre-trained in Portuguese text, with models such as GPT-4, GPT-J and LLama, among others.

ENEM candidate microdata\footnote{\label{microdados}\url{https://www.gov.br/inep/pt-br/acesso-a-informacao/dados-abertos/microdados/enem}} has also been a source of inspiration for research aiming to understand the influence of external factors on test results. Many studies ~\cite{travitzki2014does,figueiredo2014igualdade} have reported that socioeconomic conditions are some of the main factors on both the students and school average result in the test. These same data are now used by us in an attempt to analyze how the tendencies of these different groups of students may relate to the biases of a large language model and may ultimately affect its performance.  One common way of stratifying students is the socioeconomic status (SES) metric, which combines features in the candidate microdata to enable more robust analysis~\cite{soares2023medida}. This metric is currently being used by INEP, the body responsible for ENEM, to contextualize the results of their audits~\cite{inep,inep2014}. 

In light of the immense impact and rapid spread of the newer LLMs~\cite{walsh2022everyone}, especially ChatGPT, exploring their possible biases is crucial as they may extend or amplify the prevailing views in society.  Motoki et al.~\cite{motoki2023more} investigate the political leanings of ChatGPT by employing a clever strategy of asking the chatbot to impersonate public figures. They find that the model shows systematic left leaning political bias, including in regards to Brazilian politicians such as Lula. Similarly, Feng et al.~\cite{feng2023pretraining}, observes that GPT models tended to be more libertarian, likely due to the influence of web text in the pre-training corpora. 

Nevertheless, the bias held by LLMs is not solely political. Kotek et al.~\cite{kotek2023gender} indicate that many models perpetuate gender stereotypes, while Atari et al.~\cite{atari2023humans} present evidence that ChatGPT is biased towards Western, Educated, Industrialized, Rich, Democratic (WEIRD) societies. According to their findings, these were the populations ChatGPT most closely resembled when answering the World Values Survey (WVS). Such biases, in the context of ENEM, may lead to prejudiced or incorrect answer distributions that better reflect the world views of a subset of students.

An integral part of the effort of finding how these biases may relate to the real-world is by comparing the LLMs' responses with real human data. One common path taken by researchers is to take public questionnaires, surveys and other means that divulge their results. In this vein, Dillion et al.\cite{dillion2023can} compare humans and LLMs as participants in psychological science, finding that these models' moral judgments are generally well aligned with human ones. Back in the context of the WVS, Durmus et al.~\cite{durmus2023towards} contrast human and AI responses, finding that, by default, the model responses most resemble western countries, even when prompting in non-English languages. They also find, however, that by asking the model to consider a certain country's perspective, the response gets closer to the actual population responses, in some cases reflecting a harmful stereotype instead. 

\section{Data Preparation}

\subsection{Socioeconomic Status}
\label{ssec:ses}
The Socioeconomic Status (SES) index follows methodological developments to capture families socioeconomic status of students performing proficiency tests such as the Brazilian ENEM~\cite{soares2004quality}. The index should capture parents' education, family wealth, and home educational resources. The literature on the determinants of education quality and quantity stresses the importance of
family determinants in school proficiency, in contrast to the effect of school material resources.

Coupled with academic developments towards constructing the SES index, the Brazilian Ministry of Education (MEC) research institute (INEP) adopted a methodology fitted to their proficiency tests that would capture the socioeconomic gradient. This paper replicates the suggested official methods to calculate the individual SES indices~\cite{inep}. Following this methodology, it is possible to stratify the socioeconomic status of the students participating in the ENEM exam.

In this process, it is worth noting that the features utilized were the ones common to both the ENEM questionnaire and the Basic Education Assessment System (SAEB) survey, used in the methodology. The description of those features can be seen in Appendix A\ref{appendix:nsefeatures}. Appendix D\ref{appendix:level_descriptions.} specifies category descriptions, highlighting the significance of the questionnaire answers for each level. Note that lower levels represent more vulnerable and less advantaged groups, whereas higher levels indicate wealthier groups.

\subsection{ENEM Dataset}
\label{ssec:enem-dataset}

The ENEM exam consists of four sections — Languages, Humanities, Natural Sciences, and Math — each containing 45 multiple-choice questions with 5 options (A-E). Five questions of the Languages section are dependent on the candidate's foreign language of choice — English or Spanish. Each question pertains to an ability from the official Reference Matrix \footnote{\url{https://download.inep.gov.br/download/enem/matriz_referencia.pdf}}, which allows for their grouping into 30 distinct competencies. Additionally, candidates must write an essay on a topical subject. All 180 questions are made publicly available every year, alongside the official answer key.

In every application, there are different color variants of ENEM, containing mostly the same questions but in different orders. This is done for two reasons: to avoid cheating and also to allow for different tests to be created for people with disabilities. In this work, we use the orange tests for evaluating the language models, as they are made for people with vision impairment, and, as such, all images are followed by detailed descriptions, which may help the LLMs to better answer the questions. We compile these data for 2022 to create a dataset used to prompt the models. Each entry in the dataset consists of the question, the 5 possible answers, the correct answer, the year, the subject and the language of the question (Portuguese, English or Spanish). Table \ref{tab:enem} shows an example of one such question translated to English. 

\begin{table}[ht]
\noindent\fbox{%
    \parbox{\columnwidth}{%
    \begin{center}
               \textbf{Question 55}\\
    \end{center}
       Born in Lebanon, two sisters could not be registered in the country, because it is required that children born in Lebanon be born to Lebanese fathers and mothers. Their parents, of Syrian nationality, were also unable to register them in their country of origin. In Syria, children are only registered by officially married parents, which was not their case. In situations like the one presented in the text, when people are born they are already in the socio-political condition of:\\
       A: exiled\\
       B: stateless\\
       C: fugitives\\
       D: refugees\\
       E: clandestine\\
       Correct answer: B stateless\\
       Year: 2022, 
       Subject: Humanities, 
       Language: Portuguese
       }}
    \caption{Example of a question from ENEM translated to English.}
    \label{tab:enem}
\end{table}

In addition to the answer keys, questions, and competencies, INEP, the government organization responsible for ENEM, also makes available the data from the test takers in the form of what they call the ``ENEM microdata". This includes a questionnaire about the students socioeconomic status, all their answers and grades, their gender, race, and age, as well as data about their school (city, public or private, etc). That is the information we use to compute the SES metric described in the section \textbf{Socioeconomic Status}\ref{ssec:ses}.

We compile these data in a human answer dataset, where, for every question, we calculate the answer distribution for all humans that answered that question, regardless of test color. This is made possible by the existence of a unique identifier for every distinct question. We also calculate the same distributions aggregated by socioeconomic status (SES). A summary of the performance in ENEM for the different SES levels can be seen in Table \ref{tab:enem_dataset}.

\begin{table}[bt]
    \centering
    \small
    \begin{tabular}{rrrrrr} \toprule
        \textbf{SES} & \textbf{Lang.} & \textbf{Humanities} & \textbf{Nat. Sci.} & \textbf{Math} & \textbf{All} \\ \midrule
        \textbf{I} & 0.68 & 0.66& 0.24& 0.27 & 0.47\\
        \textbf{II} & 0.70 & 0.66& 0.26&  0.31 & 0.49\\
        \textbf{III} &0.76 & 0.73& 0.28& 0.38 &0.54\\
        \textbf{IV} &0.80 & 0.73& 0.28&  0.40& 0.56\\ 
        \textbf{V} & 0.80& 0.73& 0.33& 0.43 & 0.58\\ 
        \textbf{VI} &0.86 & 0.82& 0.42& 0.56 & 0.67\\ 
        \textbf{VII} & 0.88&0.86 & 0.55& 0.59 & 0.72\\
        \textbf{VIII} & 0.90& 0.88& 0.60& 0.68 & 0.77\\
        \bottomrule
    \end{tabular}
    \caption{Average percentage of correct answers per ENEM subject by SES level. Notice how the percentage grows drastically with the increase in socioeconomic status.}
    \label{tab:enem_dataset}
\end{table}

Parallel to the multiple-choice ENEM dataset, we also compile a collection of human written 2022 ENEM essays from two sources. The first set of these is comprised of the 27 publicly available maximum grade (1000) essays written by ENEM candidates for the official exam. The second contains 34 user-submitted essays graded according to the ENEM criteria, sourced from Brasil Escola, a reputable website\footnote{https://vestibular.brasilescola.uol.com.br/banco-de-redacoes/tema-desafios-para.htm} widely used by Brazilian students. Ideally, we would have all of the essays written for the ENEM test. However, most of them are private, with only a few top-scorers being publicized each year, hence why we settled for these 61 samples with varying grades. 

The official list of metrics for ENEM essay evaluation is public information\footnote{\label{cartilha-redacoes}\url{https://www.gov.br/inep/pt-br/centrais-de-conteudo/acervo-linha-editorial/publicacoes-institucionais/avaliacoes-e-exames-da-educacao-basica/cartilha-de-redacao-do-enem-2022-participante}}. A summary of these guidelines can be found in Table \ref{tab:essay-guidelines}. In short, students must write a cohesive, well-developed and grammatically correct text on a specific political problem. They must make an educated use of intertextuality, mentioning relevant cultural artifacts to substantiate their arguments. Finally, they must conclude their essay with an intervention proposal to solve the given problem. The theme for the 2022 edition of the exam was ``Challenges in valuing traditional communities and peoples in Brazil". 

\begin{table}[bt]
    \centering
    \small
    \begin{tabularx}{\columnwidth}{cX} \toprule
        \textbf{Criterion} & \textbf{Description} \\ 
        \midrule
        \textbf{C1} & Adherence to appropriate vocabulary and grammar.\\
        \textbf{C2} & Pertinence and variety in cultural repertoire.\\
        \textbf{C3} & Coherence and effectiveness of rhetoric.\\
        \textbf{C4} & Cohesion in the linkage of ideas.\\
        \textbf{C5} & Inclusion of a proper intervention proposal.\\
        \bottomrule
    \end{tabularx}
    \caption{Summary of official criteria for essay evaluation in the ENEM exam.}
    \label{tab:essay-guidelines}
\end{table}

\subsection{Large Language Models}

When choosing which LLMs to evaluate in the ENEM dataset, we wanted to strike a balance between models that are widely used in society and models that are tailored for the Brazilian Portuguese language. For this reason, we chose GPT-3.5 turbo (henceforth GPT-3.5) and GPT-4 as the popular multilingual representatives, and  MariTalk as a Brazilian alternative to them. As the OpenAI models support the Portuguese language\footnote{\url{https://help.openai.com/en/articles/8357869-chatgpt-language-support-alpha-web}}, they should be capable of interpreting ENEM questions and essay topics. 

\subsection{Prompts for Multiple-Choice Questions}

The prompting strategies implemented in this project are designed to reflect real-world scenarios. In this context, the zero-shot method is chosen as it enables the models to utilize their inherent linguistic and contextual understanding to mostly unseen questions. Consequently, our prompt consists of just an instruction, followed by the question data. For Maritalk, we utilized an instruction in Portuguese, whereas for both GPT-4 and GPT-3.5, we opted to use the English translation of the instruction.

While using English instructions may introduce a small extent of language bias, it helps prevent the GPT models from misinterpreting its task and producing unexpected answers, which directly compromises our methodological approach. The exam questions for all models, however, remain in Portuguese, preserving the original linguistic context and still enabling an adequate assessment of the model's comprehension and accuracy. The exact text used for the prompt can be seen in Appendix B \ref{appendix:multiplechoice-prompt}.

% our prompt for MariTalk consists of just an instruction, in Portuguese, followed by the question data. For both GPT-4 and GPT-3.5, we opted to utilize the English translation of the instruction since these models did not respond well to the Portuguese version, presenting unexpected token answers and lower accuracy. The exact text used for the prompt can be seen in Appendix B \ref{appendix:multiplechoice-prompt}.

To communicate with the GPT models, we employed the Chat Completions API \footnote{\url{https://platform.openai.com/}}.
The same set of parameters were used for both of them. First, we set a seed value of 10, to ensure reproducibility, and a temperature value of 0, enforcing the model to generate the most likely answers. Furthermore, we limit the tokens generated through max\_tokens = 1, tailoring the specific need for an objective and single answer, and select the top\_logprobs value as 5, in order to match the five alternatives presented in the exam question. Lastly, the logprobs parameter was set to true, enabling the model to produce the log probabilities for the generated token.

Having the logprobs for each token is important to our methodology, as it offers a cost-effective strategy to assess the GPT models probability distributions. This process bypasses the intensive task of generating a large number of samples. Instead, the log probabilities represent the model's unnormalized outputs, allowing us to apply a softmax function to obtain a precise probability distribution for each answer choice. Furthermore, recognizing that the top 5 log probabilities might not exclusively correspond to the expected 5 answer options in all responses, we have devised a method to address this scenario. Our approach involves filtering out unexpected responses and subsequently redistributing the probabilities among the remaining valid choices. This ensures that our analysis remains standardized and reliable.

We used the API for MariTalk, with default parameters, since that is the configuration that most users of the model will encounter, except for max\_tokens = 1, which, as stated previously for the GPT models, sums up the answer. 
Since we could not directly assess MariTalk's model probability distributions, we sampled 100 answers for each question, in order to get a probability distribution, hence we set the do\_sample parameter to true. After the samples were generated, they were parsed by selecting the answers that clearly stated the alternative, either by choosing a letter between A and E, or by copying/rephrasing one of the possible options. Furthermore, responses that did not choose any alternative by the end of the text were considered as unanswered.

\subsection{Prompts for Essay Generation}

For instructing the aforementioned models on the essay task, we use the text generation endpoints using the default parameters, as we expect that would be the configuration most real users would use. Additionally, we consider both possible configurations for each of following prompting decisions, totaling four distinct prompting strategies:

\begin{itemize}
    \item \textbf{Explicit system prompting: }We analyze whether including certain instructions in accordance to the essay evaluation criteria leads to an increase in the quality of the generated texts. In particular, we ask the model to ``write an ENEM-style essay" and ``make use of cultural references, such as books and philosophers". We also list the details needed for an intervention proposal. Finally, we limit the essay to 500 words, given that students must not write more than 30 lines of text.
    \item \textbf{Inclusion of sample texts: } The official ENEM exam presents students with a few sample texts to better their understanding of the essay topic. Although students must not copy directly from them, these texts are helpful in clarifying how examiners expect candidates to approach the topic. As such, we find it relevant to test their effect on model-generated answers.
\end{itemize}

The use of varying prompts is justified on the basis that the generated essay is sensible to instructional choices \cite{louie2023prompt}. By prompting the model in the four manners described, we ensure that the resulting texts are not overly dependent on the precise instructions fed to the machine. Thus, we better emulate real interactions between LLMs and potential users.

In total, for each LLM, we generate 100 essays, split evenly among the four prompting strategies. The exact prompts are replicated in Appendix C \ref{appendix:essay-prompt}. Although our work can be expanded to a larger sample size, we choose to analyze only so many essays, as we have a limited number of human counterparts (see the section \textbf{ENEM Dataset}). This sample size, albeit relatively small, was enough for us to find statistically significant differences between the generated texts and humans.

Having generated the model responses, we opted to remove any extraneous text that did not match the structure of the ENEM essay. This includes section headers and bullet points, which were sometimes added at the end of the file. For uniformity's sake, we also chose to exclude titles, as those are optional in the official exam. The main body of the essays remained intact, and was not subject to manual or automated corrections of any type.

\section{The Multiple-Choice Tests}
In this section, we analyze the LLMs performance on the ENEM objective questions, contextualizing it with human performances and socioeconomic aspects. We start by looking into the accuracy of each of the studied LLMs in the objective tests. Then, we show the similarity between these results and the human results in the test, stratified by the different human SES groups. Finally, we contextualize how this translates to actual biases. 

%We find that, although model answers in most subjects are more similar to answers from the richest and more privileged groups, this is possibly due to the fact that these groups are more likely to answer correctly, rather than the model showing any kind of bias.

\subsection{Evaluating LLMs on ENEM}
\label{sec:evaluatingllms}

For this inspection, we consider only non-void questions from the 2022 orange variant of the ENEM exam. We assess the performance of the models over 184 questions in total (Languages = 50, Humanities = 45, Natural Sciences = 45, Math = 44). 

The three models share a common pattern: decent accuracy in Languages and a great performance in Humanities, followed by comparatively mediocre results in Natural Sciences and Math (Table \ref{tab:correct_answers}). At first glance, this distribution seems to reflect how hard each section is. On the whole, according to ENEM microdata, Brazilian students do worse in Math and Natural Sciences than in other subjects.

% Old Table for 2022

\begin{table}[ht]
    \centering
    \small
    \begin{tabular}{lrrrrr} \toprule
        \textbf{Model} & \textbf{Lang.} & \textbf{Humanities} & \textbf{Nat. Sci.} & \textbf{Math} \\ \midrule
        \textbf{GPT-3.5} & 0.76& 0.87& 0.58& 0.29\\
        \textbf{GPT-4} & 0.84& 0.93& 0.71&  0.23\\
        % \textbf{MariTalk} & 0.80& 0.96& 0.49& 0.32& 0.65\\ Old?
        \textbf{MariTalk} & 0.82 & 0.95 & 0.72 & 0.29\\
        \bottomrule
    \end{tabular}
    \caption{Accuracy by model for each exam section.}
    \label{tab:correct_answers}
\end{table}

% New Table Combining All Years - Lacking MariTalk

% \begin{table}[ht]
%     \centering
%     \small
%     \begin{tabular}{rrrrrr} \toprule
%         \textbf{Model} & \textbf{Languages} & \textbf{Humanities} & \textbf{Natural Sc.} & \textbf{Math} & \textbf{All} \\ \midrule
%         \textbf{GPT-3.5} & 0.72 & 0.86 & 0.60 & 0.25 & 0.59\\
%         \textbf{GPT-4} & 0.84 & 0.94 & 0.75 &  0.32 & 0.71\\
%         \textbf{MariTalk} & 0.82 & 0.95 & 0.72 & 0.29 & 0.68\\
%         \bottomrule
%     \end{tabular}
%     \caption{Accuracy by model for each exam section.}
%     \label{tab:correct_answers}
% \end{table}

However, it is debatable whether human-measured difficulty is the most influential factor on model performance. Rather, question structure might be more significant. Most Natural Sciences and Math items require complex, multi-step reasoning — at which LLMs appear to struggle ~\cite{stepReasoning}. In contrast, the Languages and Humanities sections mainly comprise general knowledge and interpretation tasks. As such, given their natural language processing capabilities, LLMs achieve better results in these sections.

Although LLM scores come across as human-like on the surface, it is unrealistic to claim they stem from the same variables that impact regular ENEM participants. In the following sections, we delve deeper into the differences between human and LLM behavior.

\subsection{Comparing Humans and LLMs on the Objective Tests}
\label{sec:sim}

Given the set of ENEM questions(Q) discussed previously, we compute the similarity of the answer(a) distributions from the set of students(H) with the responses from the set of models(M). We chose the Jensen-Shannon(JS) distance as the similarity metric due to the fact that it is symmetric and finite, but any other metric adequate for probability distributions could be used.

In mathematical terms, given the answers for question q defined as $a_i \in A_q=\{A,B,C,D,E,*\}$, we calculate the probabilities distributions for each question for the humans as:
            \[P_H(a|q) = \frac{n_{a,q}}{n_q}, \forall a \in A_q, q \in Q\] where $n_{q}$ denotes the total number of answers to question $q$ and $n_{a,q}$ denotes the number of specific answers to question $q$.
            
Whereas the probability for the OpenAI models is defined as:
            \[P_m(a|q), \forall a \in A_q, q \in Q, m \in M\] where $P_m(a|q)$ is the probability of model m outputting answer $a$ given question $q$.
            
For MariTalk, the calculation resembles humans, as we sample $S=100$ responses for the model before using those samples to calculate the probabilities:
    \[P_m(a|q) = \frac{n_{a,q}}{S}, \forall a \in A_q, q \in Q\]
    
Thus, the similarity over all questions, between humans and a model, is calculated as:
\[Similarity(m,H) = \frac{1}{N}\sum^N_{q=1}{JS(P_m(A_q|q),P_H(A_q|q))}\] Since we opt to use the base 2 logarithm for the Jensen-Shannon distance, it is bounded in the interval [0,1].
% marcelo, seria interessante enfatizar qual o range do 
% Jensen-Shannon distance
% definido!

Note that the set of questions may change depending on the context of an analysis. For example, if we compare the results of humans and models over the entire test, Q will be the 185 questions that comprise ENEM, meanwhile, if we decide to focus on a specific subject, Q will be the questions related to that instead. 

    \begin{figure*}[ht]
        \centering
        \begin{subfigure}[b]{0.475\textwidth}
            \centering
            \includegraphics[width=0.8\textwidth]{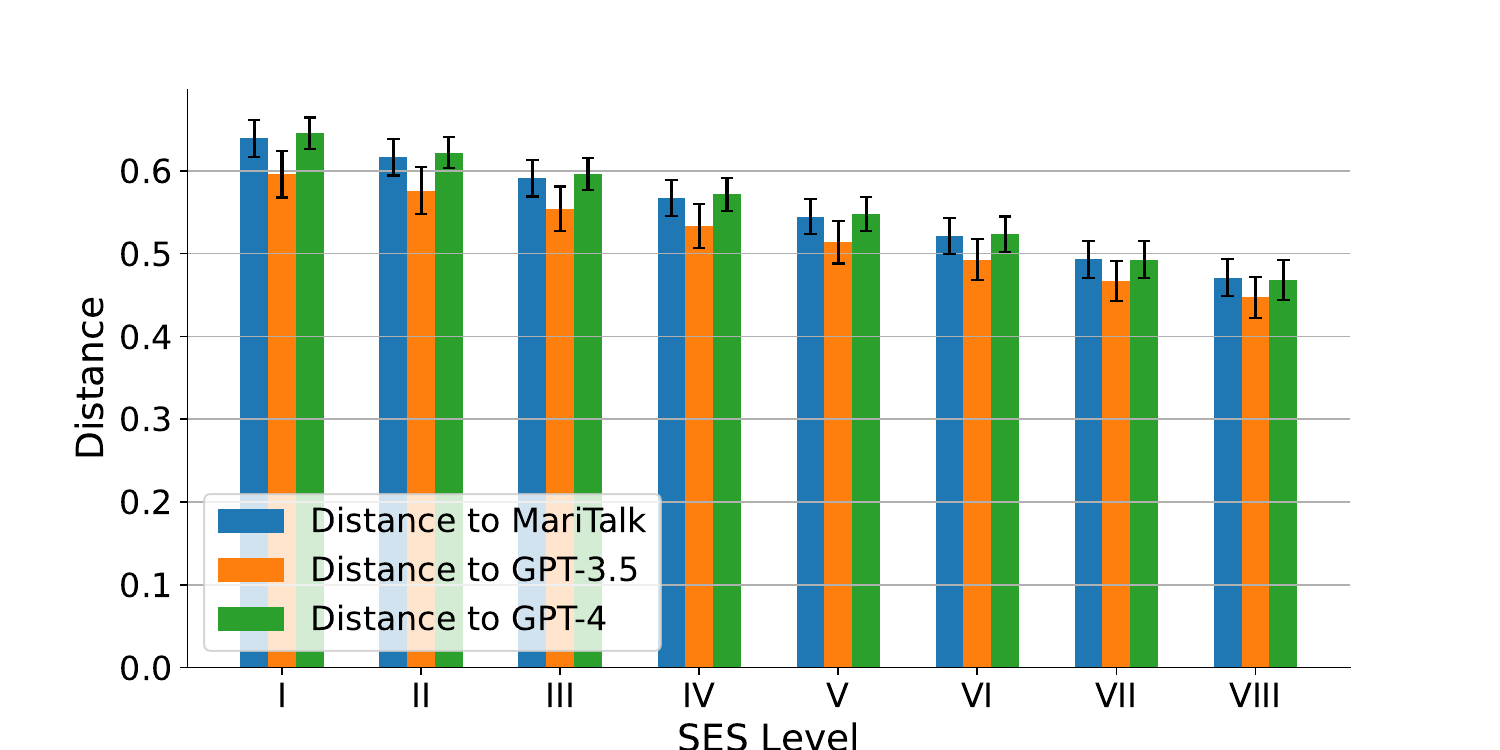}

            \caption{{\small Distance between human and models' probability distributions by SES level in the Languages ENEM test}}    
            \label{fig:ses_lc}
        \end{subfigure}
        \hfill
        \begin{subfigure}[b]{0.475\textwidth}  
            \centering 
            \includegraphics[width=0.8\textwidth]{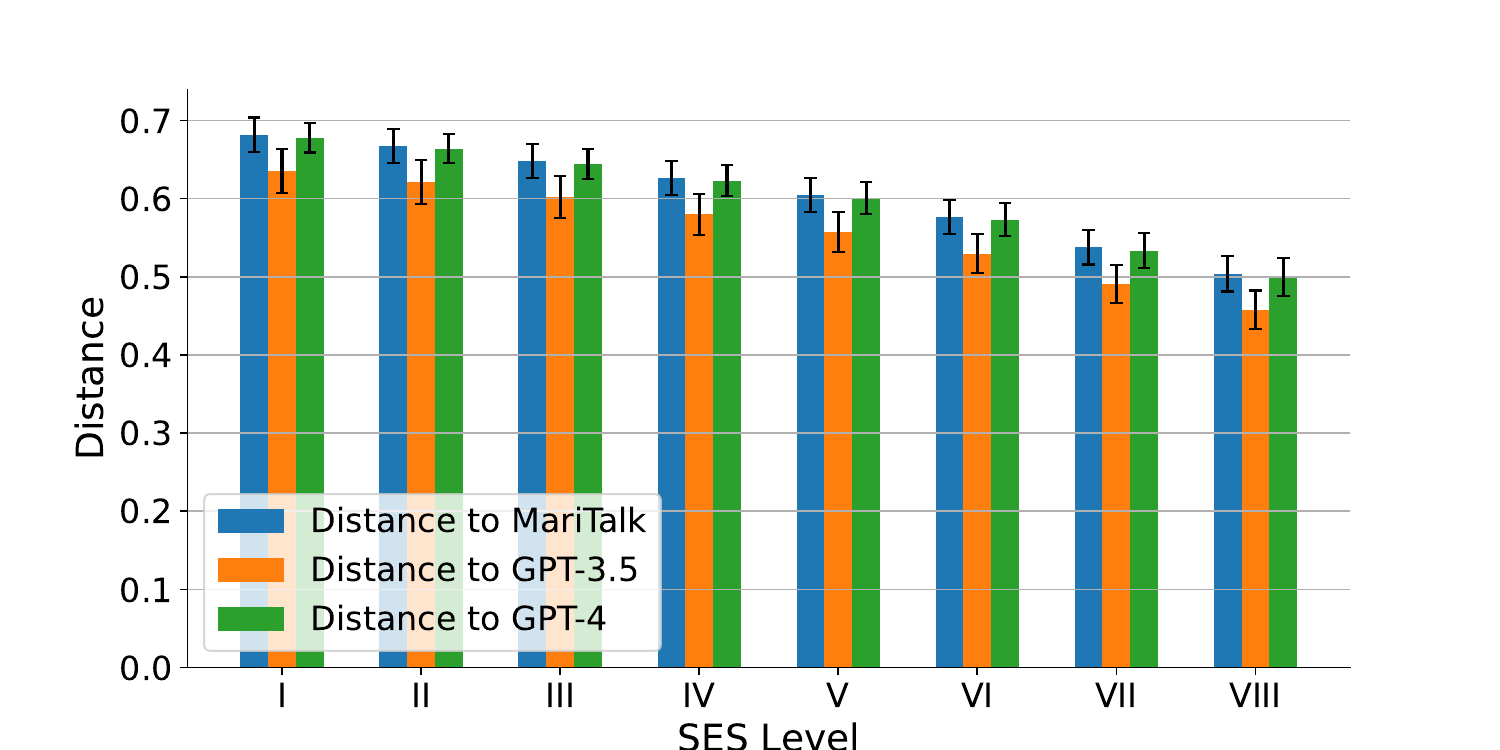}
            \caption{{\small Distance between human and models' probability distributions by SES level in the Humanities ENEM test}}    
            \label{fig:ses_ch}
        \end{subfigure}
        \vskip\baselineskip
        \begin{subfigure}[b]{0.475\textwidth}   
            \centering 
            \includegraphics[width=0.8\textwidth]{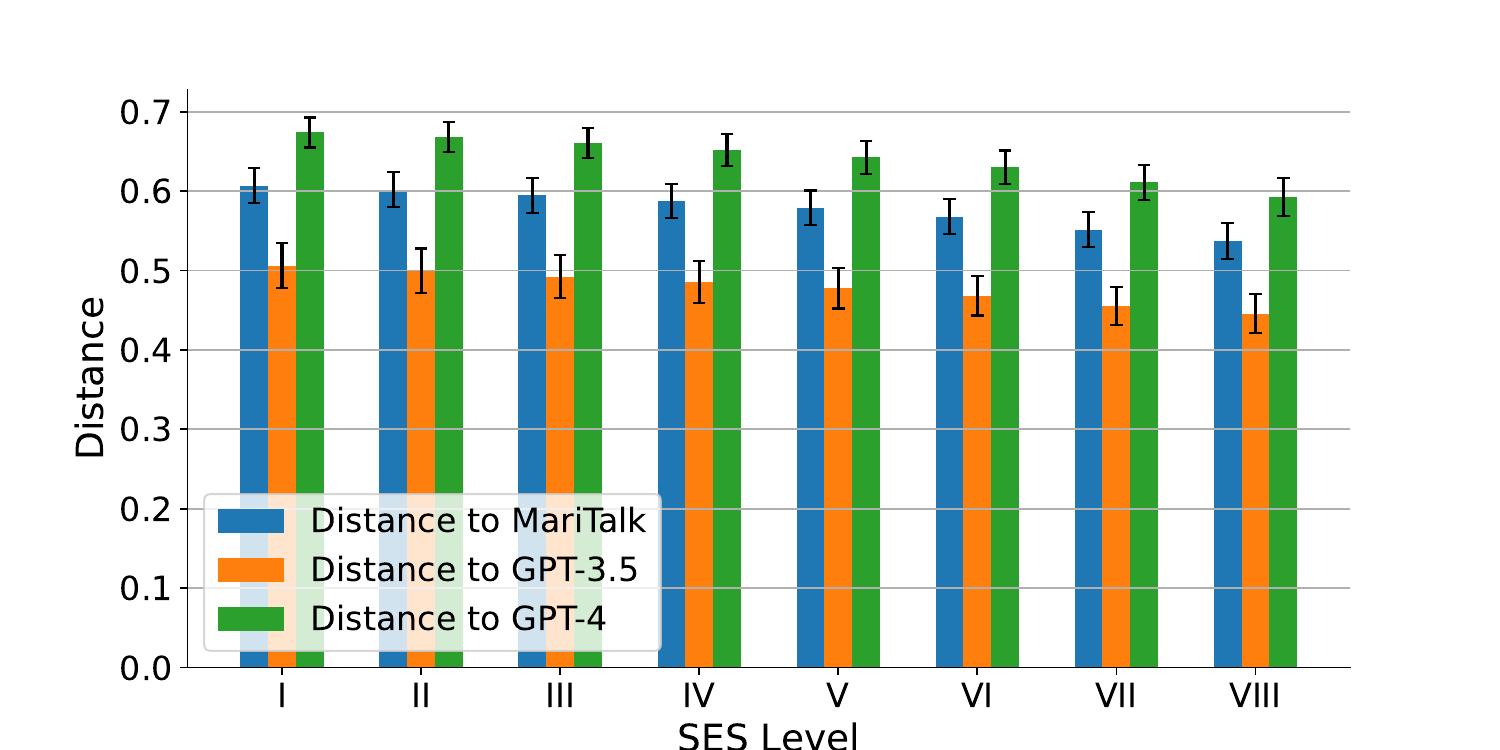}
            \caption{{\small Distance between human and models' probability distributions by SES level in the Natural Sciences ENEM test}}    
            \label{fig:ses_cn}
        \end{subfigure}
        \hfill
        \begin{subfigure}[b]{0.475\textwidth}   
            \centering 
            \includegraphics[width=0.8\textwidth]{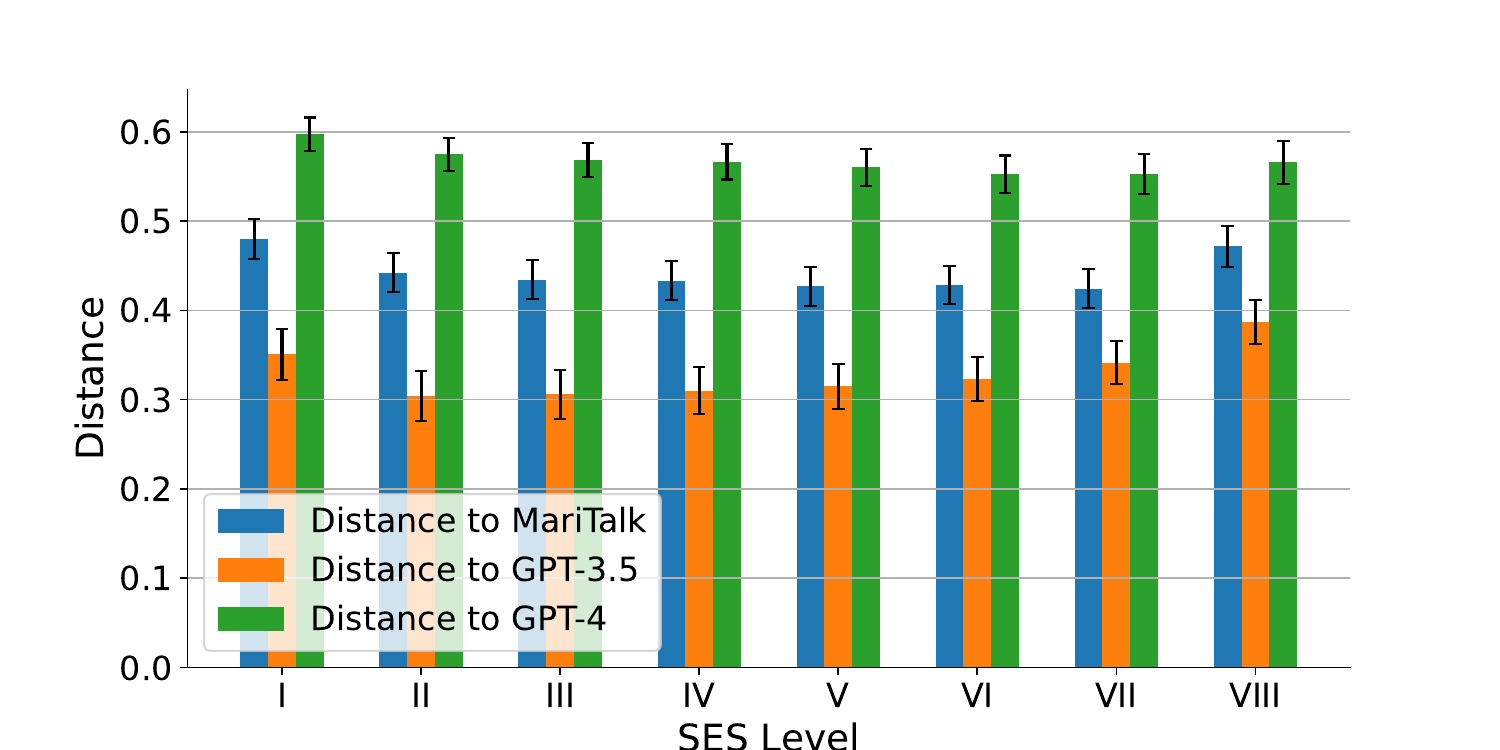}
            \caption{{\small Distance between human and models' probability distributions by SES level in the Math ENEM test}}    
            \label{fig:ses_mt}
        \end{subfigure}
        \caption{ Distance between human and models' probability distributions by SES level by ENEM test subject. The error bars represent the 95\% confidence interval.} 
        \label{fig:ses_all}
    \end{figure*}

Figure \ref{fig:ses_all} illustrates the relationship between SES levels and the JS distance between human and models' distribution. It is possible to see that for most tests, an increase in economic condition is correlated with a reduction in distance to the models, with the exception of the Math test - see Figure \ref{fig:ses_mt}. We reckon this happens due to the worse performance and lower probabilities attributed to the correct answer by the models in this test, which more closely matches the fact that most Brazilian students also do badly, especially those of lower SES. Notice how, despite this tendency, the difference between adjacent SES levels is not statistically significant, especially for the Natural Science test (Figure \ref{fig:ses_cn}).

These results could suggest that the model is biased towards groups with higher socioeconomic status. However, there is another possibility: the lower distance is determined not by bias, but rather by the probability of both the model and the socioeconomic group getting an answer correct. We will expand on that in the next section.

It is also interesting to discuss about the models, where we see that for all subjects GPT-3.5 is closest to humans. This is notable as it happens both in subjects where GPT-3.5 is beaten by all other models, as well as in subjects at which it excels. This is likely due GPT-3.5 being less confident in its response, whereas the other models tend to assign extremely high probabilities to one of the answers - see Table \ref{tab:avgprob}. Since humans are unlikely to all choose the same answer, this uncertainty makes GPT-3.5 answers the most human like, even in tests where it does worse than the average human.

\begin{table}
    \centering
    \begin{tabular}{lr} \toprule
        \textbf{Model} & \textbf{Probability} \\ \midrule
         GPT-3.5 & 0.747 +- 0.003\\
         GPT-4 & 0.896 +- 0.002\\
         MariTalk & 0.833 +- 0.002\\ \bottomrule
    \end{tabular}
    \caption{Average probabilities of the alternative most likely to be chosen by the LLM in the multiple-choice questions.}
    \label{tab:avgprob}
\end{table}

\subsection{Isolating the Effect of SES}
\label{sec:reg}

In the previous section we observe that, for most subjects, the distance between human and model answer distributions shrink as the SES level grows. However, it remains unclear whether this is due to an intrinsic socioeconomic bias of the large language models or if it is simply a result of these wealthier levels demonstrating higher accuracy, which matches the high accuracy of the studied models. This relationship between socioeconomic status and human standardized test results has long been noted in previous research, such as \cite{zwick2002sat} in the US and \cite{travitzki2014desigualdades} in Brazil.

To isolate the effect, it is desirable to model the overlap between a model answer probabilities and each human answers given the accuracy of the human and their SES score. We define this overlap for each human as:
    \[O(h,m) = \sum^N_{q=1}\frac{P_m(A_q = a_{q,h})}{N} \] where $a_{q,h}$ is the human response to question q, while $P_m(A_q = a_{q,h})$ is the probability of model answering the same as the human. This metric measures the average probability of a model generating the same answer of a given human.

\begin{figure}[tb]
    \centering
            \includegraphics[width=0.75\linewidth]{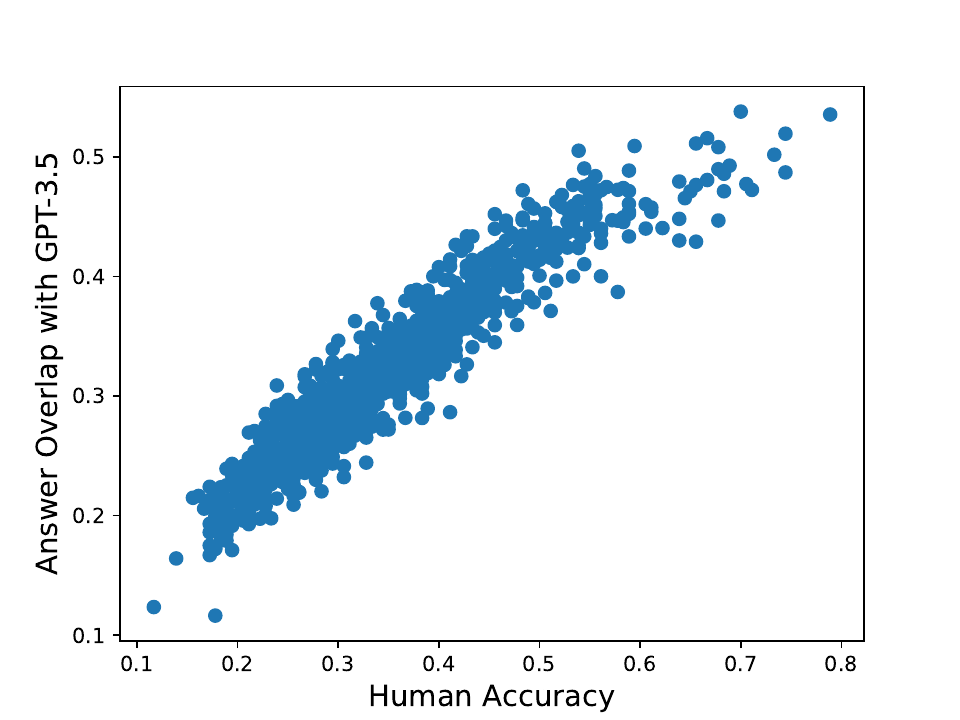}
            \caption{Relationship between human accuracy and the overlap with GPT-3.5 answers. Notice how the relationship is almost linear.}
            \label{fig:scatter_acc}
\end{figure}

\begin{figure}[tb]
    \centering
            \centering
            \includegraphics[width=0.75\linewidth]{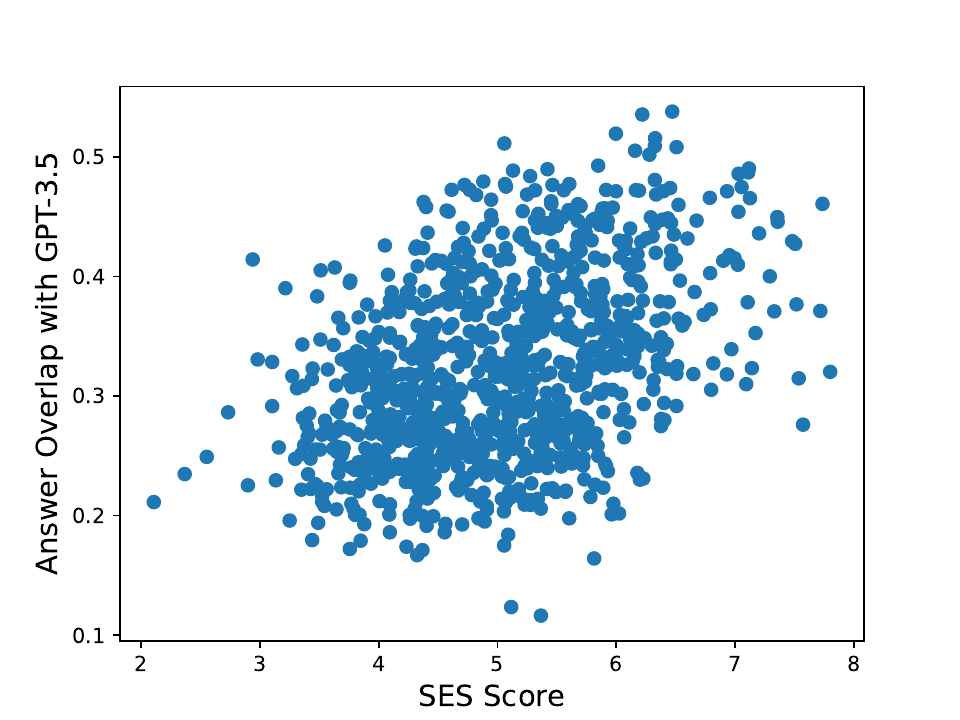}
            \caption{Relationship between human SES Score and the overlap with GPT-3.5 answers.}
            \label{fig:scatter_ses}
\end{figure}

Figures \ref{fig:scatter_acc} and \ref{fig:scatter_ses} show the relationship between human accuracy and overlap, and SES score and overlap, respectively, for the GPT-3.5 model. It is evident from the figures that the overlap is highly correlated (Pearson’s correlation = 0.94, p$<$0.001) with human accuracy, although the effect of SES score is not as clear, despite moderate correlation (Pearson’s correlation = 0.41, p$<$0.001). It is important to note that human accuracy and SES score are only moderately correlated (Pearson’s correlation = 0.44, p$<$0.001), with the amount to which they correlate decreasing as SES gets higher, thus, human accuracy is not a proxy for SES.

% \begin{figure*}[ht]
%         \centering
%         \begin{subfigure}[b]{0.475\textwidth}
%             \centering
%             \includegraphics[width=0.8\linewidth]{Figures/accxoverlap.pdf}
%             \caption{\small{Relationship between human accuracy and the overlap with GPT-3.5 answers. Notice how the relationship is almost linear.}}
%             \label{fig:scatter_acc}
%         \end{subfigure}
%         \hfill
%         \begin{subfigure}[b]{0.475\textwidth}  
%             \centering
%             \includegraphics[width=0.8\linewidth]{Figures/nsexoverlap.pdf}
%             \caption{\small{Relationship between human SES Score and the overlap with GPT-3.5 answers.}}
%             \label{fig:scatter_ses}
%         \end{subfigure}
%     \caption{Scatter plots of a sample of 1k humans, showing the relationship between different features and the overlap ($O$) in answers with GPT-3.5.}
%     \label{fig:relation}
% \end{figure*}
    
We opt to use a linear regression to model the effect of both human accuracy as well as SES score on the overlap. Table \ref{tab:regression} shows the results for the multiple regressions for each LLM. We also investigate linear regressions for the GPT-3.5 using individual features, aiming to ascertain the impact of each predictor when isolated (Table \ref{tab:regression_gpt3}).

\begin{table}[tb]
\centering
\small
\begin{tabular}{llrcc} \toprule
Model    & Feature        & Coef.                & R2    & Adj. R2 \\ \midrule
GPT-3.5  & Constant       & 0.1024       & 0.889 & 0.889       \\
         & Human Acc. & 0.6300       &       &             \\
         & SES Score      & -0.0006      &       &             \\ \hline
GPT-4    & Constant       & 0.0779       & 0.904 & 0.904       \\
         & Human Acc. & 0.7539       &       &             \\
         & SES Score      & -0.0008       &       &             \\ \hline
MariTalk & Constant       & 0.0098       & 0.893 & 0.893       \\
         & Human Acc. & 0.6535       &       &             \\
         & SES Score      & -0.0006      &       &      \\ \bottomrule      
\end{tabular}
\caption{Results of the multiple regression models for predicting human answer overlap with LLM by using SES score and human accuracy as predictors. All p-values are \textless0.001}
\label{tab:regression}
\end{table}

\begin{table}[bt]
\centering
\small
\begin{tabular}{lllll} \toprule

Model          & Feature        & Coef.                & R2    & Adj. R2 \\ \midrule
Multivariate   & Constant       & 0.1024      &  0.889 & 0.889       \\
               & Human Acc.& 0.6300      &        &             \\
               & SES Score      & -0.0006     &        &             \\ \hline
SES only       & Constant       & 0.1642      &  0.174 & 0.174       \\
               & SES Score      & 0.0321      &        &             \\ \hline
Human Acc.  & Constant       & 0.1003      &  0.889 & 0.889       \\
        only       & Human Acc. & 0.6277      & &            \\ \bottomrule
\end{tabular}
\caption{Results of linear regression analysis isolating individual features to predict human answer overlap with GPT-3.5 answers. All p-values are \textless0.001.}
\label{tab:regression_gpt3}
\end{table}

We find that, although both human accuracy and SES score are statistically significant predictors for the overlap, most of the variance can be explained by using human accuracy alone. In fact, adding SES score to the model does not improve R2. Interestingly, when considering SES as the only feature, it appears with a positive coefficient (0.0321, p$<$0.001), while it has a very small negative coefficient in the multiple regression (-0.0006, p$<$0.001). The positive coefficient is possibly a consequence of the fact that a higher SES leads to higher accuracy on average. Meanwhile, the negative coefficient means that for a given value of human accuracy, the model responses are on average slightly more similar to humans of lower socioeconomic status, which suggests that the model does not show bias towards the more privileged groups on the multiple-choice tests. The aforementioned tendencies appear for all models, but the results for MariTalk and GPT-4 have been omitted for brevity.

\section{The ENEM Essay}

In this section, we compare LLM-generated text to real human essays from two sources: publicly available maximum grade ENEM texts and ENEM-style essays written by students for training purposes. Our goal is to determine how different text produced by machines is from that produced by real people. We first contrast syntactic metrics from the LLM essays with the human data, aiming to find structural differences used by either group in the Portuguese language. Then, we study whether the choice of words of the groups are distinct enough to allow for separation between them.

\subsection{Linguistic Differences Between Machine-Generated and Human Text}
Metrics of syntactic complexity are a multidimensional basis for the understanding of differences in writing processes~\cite{9fa846f5114d4ed4b19925446113a794}. Syntactic complexity refers to the range and sophistication of forms that appear in language production~\cite{ai2013corpus}. Historically, these metrics have been employed to assess the writing proficiency of students learning a second language, given the existence of a baseline against which to gauge their performance. ~\cite{wolfe1998second} 

In this section, we employ a selection of them to compare the writing abilities of native Portuguese students who have taken the ENEM writing test, officially or otherwise, with LLMs. As a reminder, for this analysis, we consider the 61 human samples described in section \textbf{ENEM Dataset}. In total, we use five metrics of syntactic complexity~\cite{ai2013corpus} obtained from the output of the UDPipe~\cite{udpipe:2017} software with the Portuguese-Bosque model. Using them, it was possible to analytically describe specific elements of texts and then compare how different types of writing resemble each other.

Some of the metrics depend on the definition of certain terms. For instance, sentences represent complete syntactic structures, comprising word sequences that articulate entire expressions, questions, or commands. Within sentences, clauses form cohesive units, containing subjects and predicates to construct complete statements. Dependent clauses rely on independent clauses for grammatical coherence, showcasing the multifaceted nature of sentence structure. T-units, or ``thought units," encapsulate complete ideas, comprising main clauses, in addition to any associated dependent or embedded clauses ~\cite{hunt1965grammatical}. Meanwhile, coordinate phrases merge similar syntactic elements—adjectives, adverbs, nouns, or verbs—through coordinating conjunctions like ``and" or ``but", fostering cohesive expression within sentences.

Each metric provides insights into different aspects of linguistic structure and complexity. The Mean Length of Sentences (MLS) measures the average number of words per sentence, reflecting the overall sentence length and potentially indicating the level of detail or complexity in expression. Similarly, the Mean Length of Clauses (MLC) and the Mean Length of T-Unit (MLT) quantify the average number of words per clause and per T-Unit, respectively, offering an indication of production unit length. The Coordinate Phrases per Clause (CPC) metric assesses the frequency of coordinate phrases within clauses, highlighting instances where multiple syntactic elements are conjoined to convey additional information or complexity. Finally, the T-unit per Sentence (TS) metric calculates the ratio of T-units to sentences, providing insight into the structural complexity and cohesion of sentences.

We use these concepts to compare human and machine-generated essays for each LLM. The average length of these essays varies by model: GPT-3.5's essays have an average of 481 ± 13.27 words (mean ± std), GPT-4's have 532 ± 11.50 words, and MariTalk's have 382 ± 3.17 words. Conversely, human-written essays have an average length of 517 ± 40.45 words. Notice how the length of human essays has much more variance than the LLM generated text. This is interesting as, while two prompting templates (see Appendix \ref{appendix:essay-prompt}) explicitly suggest 500 words (matching the typical size of a human essay), not all of them do, which indicates that the GPT LLMs are able to implicitly match the size of average human essays, as seen by the average length and low variance. The same can't be said for MariTalk.

\begin{figure}[ht]
    \centering
    \includegraphics[width=0.9\linewidth]{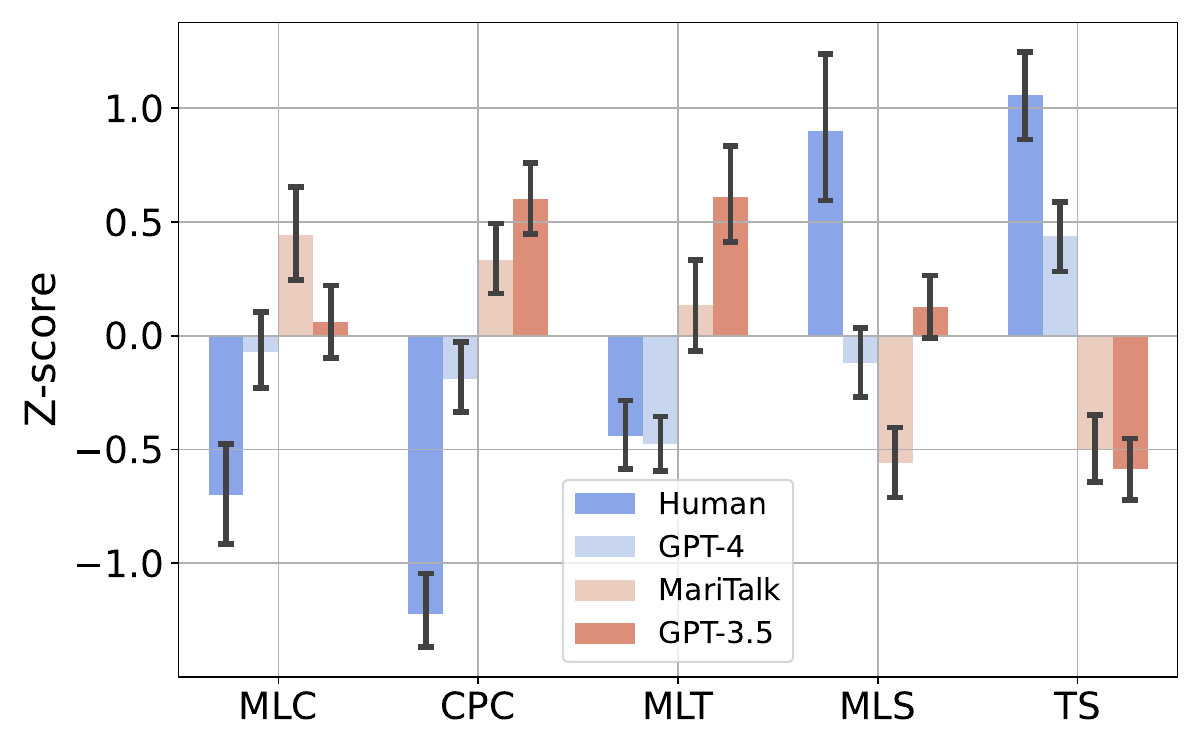}
    \caption{\small{Z-score for syntactic metrics. Notice that the GPT-4 model tends to be closer to humans than other models in most features.}}
    \label{fig:statisticald}
\end{figure}

Since we want to compare the various models with human written text, we use the aforementioned syntactic metrics to evaluate how their essays may or may not approach that of humans. Table~\ref{tab:report_values} and Figure~\ref{fig:statisticald} show, respectively the value of each metric as well as the z-scores associated with each group. We can observe that all of the proposed syntactic complexity metrics, besides MLT, show statistically significant differences when comparing human essays with LLM generated ones. 

This suggests that objective questions are not the only task on which LLMs differ from humans in the Brazilian Portuguese language, as we see similar tendencies in the subjective writing section of ENEM. It is also possible to observe that humans tend to compose phrases using a larger quantity of compound sentences than machines do, as evidenced by their higher TS and lower MLC. This higher TS comes as a consequence of humans opting for shorter clauses, resulting in longer sentences with more clauses, as evidenced by the MLS metric.

Furthermore, we can see that, unlike in the multiple choice test, in the essay, GPT-4 is closer to humans across all metrics, which could be due to its greater understanding of natural language as a larger model. Interestingly, in spite of being pre-trained on Brazilian Portuguese, MariTalk is the farthest model from Brazilian students for many of the syntactic metrics.

% As we are interested in discovering the differences that the various models show in relation to humans, we employed a statistical approach to validate whether two groups exhibited significant differences using t-test. Figure \ref{fig:statisticald} illustrates the metrics in which humans differ from all LLMs. As can be seen, all the metrics are of the syntactic complexity type. It can be observed that humans tend to compose phrases using more compound sentences than machines, as evidenced by their higher TS (T-unit by sentence ratio), and lower MLC (mean length of clause). Aside from the GPT-4 model, the average length of a essay (measured in sentences) is approximately the same for humans and LLMs. In order to achieve a higher level of TS, humans often opt for shorter clauses, resulting in longer sentences with more clauses, despite the reduced clause size, as evidenced by the MLS metric. 

% This indicates that humans write clauses with fewer words, to be able to write more compound sentences per phrase.

% \begin{table*}[bt]
% \centering
%     \begin{tabular}{lrrrrr}
%     \toprule
%      \textbf{class} & \textbf{MLC} & \textbf{CPC} & \textbf{MLT}  & \textbf{MLS} & \textbf{TS} \\

%     \midrule
% GPT-3.5 & 22.03 ± 0.90 & 3.02 ± 0.16 & 92.16 ± 6.74 & 34.50 ± 1.00 & 0.41 ± 0.03 \\
% GPT-4 & 21.34 ± 0.91 & 2.24 ± 0.15 & 58.01 ± 3.68 & 32.74 ± 1.08 & 0.60 ± 0.03 \\
% Human & 18.02 ± 1.17 & 1.22 ± 0.16 & 59.13 ± 4.94 & 40.05 ± 2.39 & 0.71 ± 0.04 \\
% MariTalk & 24.06 ± 1.09 & 2.76 ± 0.15 & 77.19 ± 6.50 & 29.59 ± 1.11 & 0.43 ± 0.03 \\
%     \bottomrule
%     \end{tabular}
%     \caption{Value of syntactic metrics which show statistically significant difference between groups. It can be seen that values for the human group are very distant from the others.}
% \label{tab:report_values}
% \end{table*}

\begin{table}[bt]
\centering
    \small
    \begin{tabular}{lrrrrr}
    
    \toprule

     \textbf{Model} & \textbf{MLC} & \textbf{CPC} & \textbf{MLT}  & \textbf{MLS} & \textbf{TS} \\

    \midrule
GPT-3.5 & 22.03 & 3.02 & 92.16 & 34.50 & 0.41 \\
GPT-4 & 21.34 & 2.24& 58.01 & 32.74& 0.60 \\
Human & 18.02 & 1.22 & 59.13 & 40.05 & 0.71 \\
MariTalk & 24.06 & 2.76 & 77.19 & 29.59  & 0.43 \\
    \bottomrule
    \end{tabular}
    \caption{Value of syntactic metrics for humans and language models. It can be seen that values for the human group are very distant from the others. It is important to note that the differences for MLT are not statistically significant.}
\label{tab:report_values}
\end{table}

\subsection{Comparing Human and LLM on ENEM-Style Essays}

Inspired by the text comparisons between different LLM described demographic groups in \cite{cheng2023marked}, we use a one-vs-all SVM to distinguish between the four different categories of ENEM essays: the human essays, as well as those of the LLMs being tested (GPT-3.5, GPT-4 and MariTalk). 

For this analysis, we remove stopwords and represent each text as a bag-of-words representation, which is highly interpretable and allows us to identify which words contributed most heavily for the classification. We split each category into 70\% train and 30\% test data.

We find that human and LLM essays are easily separable, as we are able to achieve mean accuracy of 0.98 ± 0.02 (mean ± std), lending strength to our previous observation that human and LLM answers are not all that similar. Additionally, this also suggests that texts produced by different models are also distinct. This becomes even more evident when looking at the top words for the classification of each group (Table \ref{tab:topwords}), as we see that while human top words relate to society and nature, with words such as country, government, and involvement being most discriminative, GPT-3.5 seems to focus instead on actions, e.g. build, participation, and strengthening. GPT-4 seems to focus on culture, with words such as cultures, philosopher, traditions, and book being among the most relevant, showing the capability of this model to adapt to the need to include cultural repertoire in ENEM-style essays. Finally, MariTalk takes a different approach, with top words focusing on diversity and disputes, including fight, confront, plurality, recognition and marginalization. The dominance of these terms communicates a firmer adherence to the essay's theme of empowering the traditional peoples of Brazil.

\begin{table}[ht]
\centering
\small
\begin{tabular}{c|p{6.3cm}}
\hline
\textbf{Group} & \textbf{Translated Words} \\ \hline
Humans & Country, natural, part, shape, involvement, government, resources, access, threat, devaluation.\\ \hline
GPT-3.5 & Fundamental, society, strengthening, fair, guarantee, groups, build, faced, solely, participation.\\ \hline   
GPT-4 & Cultures, necessity, traditions, cultural, people, therefore, yet, philosopher, book, land.\\ \hline
MariTalk & Plurality, confront, fight, practice, question, crucial, existence, marginalization, recognition, maintenance, formation.\\ \hline
\end{tabular}
\caption{Top words for each group of generated essays. Notice the stark difference between human and LLM words. The words were translated from Portuguese to English.}
\label{tab:topwords}
\end{table}

\section{Discussion}

As large language models become increasingly widespread in society, the question of whom they represent is pivotal in understanding many of the problems that may arise from their usage. Although much has been done previously in the English language, with a plethora of psychological, social and political questionnaires being applied to the models, far less has been done in Brazilian Portuguese.

Many studies focus on how the large language models may display capabilities and generate answers similar to those of humans on certain conditions. However, very few studies focus instead on how these models differ from humans. In this work, we presented a case study using the 2022 ENEM test as a base to show a situation where human and LLM responses are significantly different, serving as a contrast to previous work which attempted to study such models as if they were human~\cite{atari2023humans,pan2023llms}.

While we initially sought to explore LLMs' socioeconomic biases by contrasting their outputs with humans responses made public by the large ENEM microdata, during our research, we found no significant biases of this kind, with similarity between model and human responses being explained by other factors. This is true for all studied models, regardless of being pretrained on multilingual (GPT-3.5, GPT-4) or Portuguese (MariTalk) data.

For instance, at first, when comparing human and LLM answers in the multiple-choice test, we observed a reduction in distance as the human group gets richer and more highly educated for all tests except math. However, by further inspecting these results, we find that the confounding variable, human accuracy, is most likely responsible for that decrease, so the higher similarity with the richest groups might not be related to any kind of bias, but rather to the fact that higher SES level humans tend to get more questions correct.  By fitting the average probability of a model generating the same answer of a given human ($O$) using a linear regression, we find that human accuracy alone is enough to explain most of the variance of the data.

Even though we did not identify any social or economic biases in the multiple-choice responses, we did find situations that lead the LLM responses to be closer to humans, regardless of socioeconomic factors. In this case, they are related to the confidence of a model in its answers. This is exemplified by GPT-3.5 having the closest answer distributions to humans despite being outperformed by the other models in most of the subjects. This is likely due to the fact that it tends to assign a higher mass of probability to alternatives other than the one that was predicted, which more closely matches the behavior of a group of humans, as it is highly improbable that close to 90\% of candidates choose the same alternative. Another example of this are the lower distances between humans and all three studied models being observed in the Math test --- the subject that they performed worse than all but the lowest SES level.

We find analogous results when looking at the essays, with the LLM generated text presenting syntactic properties distinct from humans. Namely, they tend to write longer clauses and shorter sentences, with less T-units per sentence. They also use more coordinate phrases per clause. This could imply that, when compared to humans, LLM essays are composed by more concise phrases, with each serving a distinct mechanical purpose. In contrast, human writing often employs different structures in the same phrase to enhance the expression of ideas, resulting in a more natural and dense writing style. In fact, the texts produced by all models and the humans are so different that a simple bag-of-words SVM model is enough to linearly separate the essays, with top words for each group being related to distinct aspects of ``traditional communities" in Brazil.

These results support the view that, although large language models have remarkable knowledge of not only language, but also a variety of subjects, they are, in essence, very different from humans, with performance that is inconsistent with what is expected of actual people, at least in Brazilian Portuguese. Similar inconsistency has been seen in other contexts, as shown by Mahowald et al.~\cite{mahowald2024dissociating}, indicating that LLMs still lack capability in certain linguistic tasks expected of humans. Future work in this area should strive to study language models not only by looking at how they may approach humans in certain conditions, but also at how they may differ. 

Another avenue to explore is whether the same tendencies observed in this work are also present in standardized tests from other countries. In theory, the methods used in this paper should be extendable to other languages with minimal changes. For the essay analysis specifically, researchers may need to take heed of the differences between languages, especially when choosing the syntactic complexity metrics.

\subsection{Limitations}

In this paper we, unfortunately, did not have the means to ask human examiners to evaluate the generated essays. Thus, qualitatively studying the ENEM essays and drawing comparisons between human and LLMs for each competency students are expected to have will be left to future work. This kind of study could be very insightful in regards to the linguistic and cognitive inconsistencies this type of model has been shown to have. 

As LLMs continue to be updated, their knowledge cutoff dates come ever closer to the present day — hence why we resorted to use only the latest available version of ENEM with microdata available (November 2022) at the time we started our research (INEP has since released the microdata for ENEM 2023 in Apr. 2024). Since GPT-4 has access to information from January 2022, assessing its abilities on older exams would be unreasonable. Moreover, this issue affects future research on this topic, as replicating our results will prove challenging once the current models become obsolete.

Another factor to consider is the language barrier. Foreign language questions notwithstanding, we conducted our research using Portuguese-language items and essay topics. As discussed by Etxaniz et al. ~\cite{thinkEnglish}, LLMs do not make use of their full multilingual aptitude unless directly prompted in English. We attempted to overcome this hurdle by translating our instructions. As detailed in the section \textbf{Prompts for multiple-choice questions}, this optimization did, in fact, lead to a boost in accuracy. 

%\clearpage
\section{Ethical Considerations Statement} 

The authors mitigate ethic concerns by solely using anonymous student data made available by the Brazilian Government and not using any identifying information when comparing the human written text. Additionally, the AI generated content complies with the ToS for the tools used.
    
\section{Positionality Statement} 

The social circumstances of the authors had a significant impact on the conclusions reached in this paper. As members of Brazilian academia, they have personal experience with the ENEM exam, which has been a pillar of university admissions since 2004. As a matter of fact, seven of the authors got accepted to a post-secondary educational institution via a selection process largely dependent on ENEM scores. Thus, they tackle this subject from a privileged perspective, in the sense that they benefited from the current ENEM exam structure. 

Additionally, the authors mostly come from a Computer Science background and, as such, are biased in their analysis of sensitive socioeconomic issues. For this reason, they sought to collaborate with an expert in the field of demography, who offered meaningful insight on handling these topics.

On the same note, the linguistic analysis was supervised by an experienced researcher in the field, as a way to ensure the appropriateness of the metrics chosen.

\section{Adverse Impact Statement}

Although this work aims to question the biases inherent to large language models, readers could misinterpret the results as a criticism of the ENEM examination system. Thus, it is relevant to emphasize that the performance of LLMs is not a reliable metric for dictating the applicability of a test in a human educational assessment context. This type of analysis escapes the scope of this study.

On a similar note, ill-intentioned individuals might misuse the variance of ENEM scores between socioeconomic groups to substantiate prejudiced rhetoric. In this respect, it must be made clear that gaps in ENEM scores are a reflection of unequal access to quality education in Brazil, among other problems commonly associated with income disparity in developing countries. Please refer to works such as ~\cite{EducacaoDesigualdade1} and  ~\cite{EducacaoDesigualdade2} for more information.

%%
%% The next two lines define the bibliography style to be used, and
%% the bibliography file.

\bibliography{bibliography}

\begin{thebibliography}{39}
\providecommand{\natexlab}[1]{#1}

\bibitem[{{ Instituto Nacional de Estudos e Pesquisas Educacionais Anísio Teixeira | INEP}(2014)}]{inep2014}
{ Instituto Nacional de Estudos e Pesquisas Educacionais Anísio Teixeira | INEP}. 2014.
\newblock Nota Técnica - Indicador de Nível Socioeconômico (Inse) das Escolas.
\newblock \url{https://download.inep.gov.br/informacoes_estatisticas/indicadores_educacionais/2011_2013/nivel_socioeconomico/nota_tecnica_indicador_nivel_socioeconomico.pdf}.
\newblock Accessed: 2024-01-17.

\bibitem[{{ Instituto Nacional de Estudos e Pesquisas Educacionais Anísio Teixeira | INEP}(2023)}]{inep}
{ Instituto Nacional de Estudos e Pesquisas Educacionais Anísio Teixeira | INEP}. 2023.
\newblock Indicador de Nível Socioeconômico do Saeb 2021 - Nota Técnica.
\newblock \url{https://www.gov.br/inep/pt-br/centrais-de-conteudo/acervo-linha-editorial/publicacoes-institucionais/avaliacoes-e-exames-da-educacao-basica/saeb-2021-indicador-de-nivel-socioeconomico-do-saeb-2021-nota-tecnica}.
\newblock Accessed: 2024-01-15.

\bibitem[{Ai and Lu(2013)}]{ai2013corpus}
Ai, H.; and Lu, X. 2013.
\newblock A corpus-based comparison of syntactic complexity in NNS and NS university students’ writing.
\newblock \emph{Studies in Corpus Linguistics}.

\bibitem[{Atari et~al.(2023)Atari, Xue, Park, Blasi, and Henrich}]{atari2023humans}
Atari, M.; Xue, M.~J.; Park, P.~S.; Blasi, D.~E.; and Henrich, J. 2023.
\newblock Which Humans?

\bibitem[{Castro(2003)}]{EducacaoDesigualdade1}
Castro, J. A.~d. 2003.
\newblock Evolução e desigualdade na educação brasileira.
\newblock \emph{Educação \& Sociedade}, 30: 673--697.

\bibitem[{Chang et~al.(2024)Chang, Wang, Wang, Wu, Yang, Zhu, Chen, Yi, Wang, Wang et~al.}]{chang2023survey}
Chang, Y.; Wang, X.; Wang, J.; Wu, Y.; Yang, L.; Zhu, K.; Chen, H.; Yi, X.; Wang, C.; Wang, Y.; et~al. 2024.
\newblock A survey on evaluation of large language models.
\newblock \emph{ACM Transactions on Intelligent Systems and Technology}, 15(3): 1--45.

\bibitem[{Chen et~al.(2024)Chen, Phang, Parrish, Padmakumar, Zhao, Bowman, and Cho}]{stepReasoning}
Chen, A.; Phang, J.; Parrish, A.; Padmakumar, V.; Zhao, C.; Bowman, S.~R.; and Cho, K. 2024.
\newblock Two Failures of Self-Consistency in the Multi-Step Reasoning of {LLM}s.
\newblock \emph{Transactions on Machine Learning Research}.

\bibitem[{Cheng, Durmus, and Jurafsky(2023)}]{cheng2023marked}
Cheng, M.; Durmus, E.; and Jurafsky, D. 2023.
\newblock Marked Personas: Using Natural Language Prompts to Measure Stereotypes in Language Models.
\newblock In Rogers, A.; Boyd-Graber, J.; and Okazaki, N., eds., \emph{Proceedings of the 61st Annual Meeting of the Association for Computational Linguistics (Volume 1: Long Papers)}, 1504--1532. Toronto, Canada: Association for Computational Linguistics.

\bibitem[{Dillion et~al.(2023)Dillion, Tandon, Gu, and Gray}]{dillion2023can}
Dillion, D.; Tandon, N.; Gu, Y.; and Gray, K. 2023.
\newblock Can AI language models replace human participants?
\newblock \emph{Trends in Cognitive Sciences}.

\bibitem[{Durmus et~al.(2023)Durmus, Nyugen, Liao, Schiefer, Askell, Bakhtin, Chen, Hatfield-Dodds, Hernandez, Joseph et~al.}]{durmus2023towards}
Durmus, E.; Nyugen, K.; Liao, T.~I.; Schiefer, N.; Askell, A.; Bakhtin, A.; Chen, C.; Hatfield-Dodds, Z.; Hernandez, D.; Joseph, N.; et~al. 2023.
\newblock Towards measuring the representation of subjective global opinions in language models.
\newblock \emph{arXiv preprint arXiv:2306.16388}.

\bibitem[{Etxaniz et~al.(2023)Etxaniz, Azkune, Soroa, de~Lacalle, Artexte et~al.}]{thinkEnglish}
Etxaniz, J.; Azkune, G.; Soroa, A.; de~Lacalle, O.~L.; Artexte, M.; et~al. 2023.
\newblock Do Multilingual Language Models Think Better in English?
\newblock \emph{arXiv preprint arXiv:2308.01223}.

\bibitem[{Feng et~al.(2023)Feng, Park, Liu, and Tsvetkov}]{feng2023pretraining}
Feng, S.; Park, C.~Y.; Liu, Y.; and Tsvetkov, Y. 2023.
\newblock From Pretraining Data to Language Models to Downstream Tasks: Tracking the Trails of Political Biases Leading to Unfair {NLP} Models.
\newblock In Rogers, A.; Boyd-Graber, J.; and Okazaki, N., eds., \emph{Proceedings of the 61st Annual Meeting of the Association for Computational Linguistics (Volume 1: Long Papers)}, 11737--11762. Toronto, Canada: Association for Computational Linguistics.

\bibitem[{Figueir{\^e}do, Nogueira, and Santana(2014)}]{figueiredo2014igualdade}
Figueir{\^e}do, E.; Nogueira, L.; and Santana, F.~L. 2014.
\newblock Igualdade de oportunidades: Analisando o papel das circunst{\^a}ncias no desempenho do ENEM.
\newblock \emph{Revista Brasileira de Economia}, 68: 373--392.

\bibitem[{Giannos, Delardas et~al.(2023)}]{giannos2023performance}
Giannos, P.; Delardas, O.; et~al. 2023.
\newblock Performance of ChatGPT on UK standardized admission tests: insights from the BMAT, TMUA, LNAT, and TSA examinations.
\newblock \emph{JMIR Medical Education}, 9(1): e47737.

\bibitem[{Giray(2023)}]{louie2023prompt}
Giray, L. 2023.
\newblock Prompt engineering with ChatGPT: a guide for academic writers.
\newblock \emph{Annals of biomedical engineering}, 15(12): 2629--2633.

\bibitem[{Gurnee and Tegmark(2024)}]{gurnee2023language}
Gurnee, W.; and Tegmark, M. 2024.
\newblock Language Models Represent Space and Time.
\newblock In \emph{The Twelfth International Conference on Learning Representations}.

\bibitem[{Hendrycks et~al.(2021)Hendrycks, Burns, Basart, Zou, Mazeika, Song, and Steinhardt}]{hendrycks2020measuring}
Hendrycks, D.; Burns, C.; Basart, S.; Zou, A.; Mazeika, M.; Song, D.; and Steinhardt, J. 2021.
\newblock Measuring Massive Multitask Language Understanding.
\newblock In \emph{International Conference on Learning Representations}.

\bibitem[{Hunt(1965)}]{hunt1965grammatical}
Hunt, K. 1965.
\newblock \emph{Grammatical Structures Written at Three Grade Levels}.
\newblock NCTE research report. National Council of Teachers of English.
\newblock ISBN 9789998777453.

\bibitem[{IBGE(2021)}]{IBGE}
IBGE. 2021.
\newblock {Desigualdades Sociais por Cor ou Raça no Brasil}.
\newblock \url{https://www.ibge.gov.br/estatisticas/sociais/populacao/25844-desigualdades-sociais-por-cor-ou-raca.html}.
\newblock Accessed: 2024-01-21.

\bibitem[{Kotek, Dockum, and Sun(2023)}]{kotek2023gender}
Kotek, H.; Dockum, R.; and Sun, D. 2023.
\newblock Gender bias and stereotypes in Large Language Models.
\newblock In \emph{Proceedings of The ACM Collective Intelligence Conference}, 12--24.

\bibitem[{Leão(2006)}]{EducacaoDesigualdade2}
Leão, G. M.~P. 2006.
\newblock Experiências da desigualdade: os sentidos da escolarização elaborados por jovens pobres.
\newblock \emph{Educação e pesquisa}, 32(1): 31--48.

\bibitem[{Liang et~al.(2022)Liang, Bommasani, Lee, Tsipras, Soylu, Yasunaga, Zhang, Narayanan, Wu, Kumar et~al.}]{liang2022holistic}
Liang, P.; Bommasani, R.; Lee, T.; Tsipras, D.; Soylu, D.; Yasunaga, M.; Zhang, Y.; Narayanan, D.; Wu, Y.; Kumar, A.; et~al. 2022.
\newblock Holistic evaluation of language models.
\newblock \emph{arXiv preprint arXiv:2211.09110}.

\bibitem[{Lu(2010)}]{9fa846f5114d4ed4b19925446113a794}
Lu, X. 2010.
\newblock Automatic analysis of syntactic complexity in second language writing.
\newblock \emph{International Journal of Corpus Linguistics}, 15(4): 474--496.

\bibitem[{Mahowald et~al.(2024)Mahowald, Ivanova, Blank, Kanwisher, Tenenbaum, and Fedorenko}]{mahowald2024dissociating}
Mahowald, K.; Ivanova, A.~A.; Blank, I.~A.; Kanwisher, N.; Tenenbaum, J.~B.; and Fedorenko, E. 2024.
\newblock Dissociating language and thought in large language models.
\newblock \emph{Trends in Cognitive Sciences}.

\bibitem[{Melo et~al.(2021)Melo, de~Freitas, de~Rezende~Francisco, and Motokane}]{renda_enem}
Melo, R.~O.; de~Freitas, A.~C.; de~Rezende~Francisco, E.; and Motokane, M.~T. 2021.
\newblock Renda familiar, acesso a bolsas de estudo e nível de educação das mães estão entre os fatores que mais afetam desempenho dos alunos no Enem.
\newblock \emph{FGV EASP}.

\bibitem[{Motoki, Pinho~Neto, and Rodrigues(2023)}]{motoki2023more}
Motoki, F.; Pinho~Neto, V.; and Rodrigues, V. 2023.
\newblock More human than human: Measuring chatgpt political bias.
\newblock \emph{Available at SSRN 4372349}.

\bibitem[{Nunes et~al.(2023)Nunes, Primi, Pires, Lotufo, and Nogueira}]{nunes2023evaluating}
Nunes, D.; Primi, R.; Pires, R.; Lotufo, R.; and Nogueira, R. 2023.
\newblock Evaluating GPT-3.5 and GPT-4 Models on Brazilian University Admission Exams.
\newblock \emph{arXiv preprint arXiv:2303.17003}.

\bibitem[{Pan and Zeng(2023)}]{pan2023llms}
Pan, K.; and Zeng, Y. 2023.
\newblock Do llms possess a personality? making the mbti test an amazing evaluation for large language models.
\newblock \emph{arXiv preprint arXiv:2307.16180}.

\bibitem[{Pavlick(2023)}]{pavlick2023symbols}
Pavlick, E. 2023.
\newblock Symbols and grounding in large language models.
\newblock \emph{Philosophical Transactions of the Royal Society A}, 381(2251): 20220041.

\bibitem[{Pires et~al.(2023)Pires, Abonizio, Almeida, and Nogueira}]{pires2023sabi}
Pires, R.; Abonizio, H.; Almeida, T.~S.; and Nogueira, R. 2023.
\newblock Sabiá: Portuguese large language models.
\newblock In \emph{Brazilian Conference on Intelligent Systems}, 226--240. Springer.

\bibitem[{Santurkar et~al.(2023)Santurkar, Durmus, Ladhak, Lee, Liang, and Hashimoto}]{santurkar2023whose}
Santurkar, S.; Durmus, E.; Ladhak, F.; Lee, C.; Liang, P.; and Hashimoto, T. 2023.
\newblock Whose opinions do language models reflect?
\newblock In \emph{International Conference on Machine Learning}, 29971--30004. PMLR.

\bibitem[{Soares(2004)}]{soares2004quality}
Soares, F. 2004.
\newblock Quality and equity in Brazilian basic education: facts and possibilities.
\newblock \emph{The challenges of education in Brazil}, 69--88.

\bibitem[{Soares and Alves(2023)}]{soares2023medida}
Soares, J.~F.; and Alves, M. T.~G. 2023.
\newblock UMA MEDIDA DO N{\'I}VEL SOCIOECON{\^O}MICO DAS ESCOLAS BRASILEIRAS UTILIZANDO INDICADORES PRIM{\'A}RIOS E SECUND{\'A}RIOS (A Measure of the Brazilian Schools' Socioeconomic Status Using Primary and Secondary Indicators).
\newblock \emph{Available at SSRN 4325674}.

\bibitem[{Straka and Strakov\'{a}(2017)}]{udpipe:2017}
Straka, M.; and Strakov\'{a}, J. 2017.
\newblock Tokenizing, POS Tagging, Lemmatizing and Parsing UD 2.0 with UDPipe.
\newblock In \emph{Proceedings of the CoNLL 2017 Shared Task: Multilingual Parsing from Raw Text to Universal Dependencies}, 88--99. Vancouver, Canada: Association for Computational Linguistics.

\bibitem[{Travitzki, Calero, and Boto(2014{\natexlab{a}})}]{travitzki2014does}
Travitzki, R.; Calero, J.; and Boto, C. 2014{\natexlab{a}}.
\newblock What does the national high school exam (ENEM) tell brazilian society?
\newblock \emph{Cepal Review}.

\bibitem[{Travitzki, Calero, and Boto(2014{\natexlab{b}})}]{travitzki2014desigualdades}
Travitzki, R.; Calero, J.; and Boto, P. 2014{\natexlab{b}}.
\newblock Desigualdades Educacionais e Socioeconômicas na População Brasileira Pré-Universitária: Uma Visão a Partir da Análise de Dados do ENEM.
\newblock \emph{Revista Brasileira de Educação}, 19(56): 675--693.

\bibitem[{Walsh(2022)}]{walsh2022everyone}
Walsh, T. 2022.
\newblock Everyone's having a field day with ChatGPT--but nobody knows how it actually works.
\newblock \emph{The Conversation. Preuzeto s https://theconversation. com/everyones-having-a-field-day-with-chatgpt-but-nobody-knows-how-it-actually-works-196378}.

\bibitem[{Wolfe-Quintero, Inagaki, and Kim(1998)}]{wolfe1998second}
Wolfe-Quintero, K.; Inagaki, S.; and Kim, H. 1998.
\newblock \emph{Second Language Development in Writing: Measures of Fluency, Accuracy, \& Complexity}.
\newblock National Foreign Language Center Technical Reports. Second Language Teaching \& Curriculum Center, University of Hawaii at Manoa.
\newblock ISBN 9780824820695.

\bibitem[{Zwick(2002)}]{zwick2002sat}
Zwick, R. 2002.
\newblock Is the SAT a ‘Wealth Test’?
\newblock \emph{Phi Delta Kappan}, 91(4): 307--311.

\end{thebibliography}

%%
%% If your work has an appendix, this is the place to put it.
\appendix

\section{Appendix A - NSE Features from ENEM Survey Questions}
\label{appendix:nsefeatures}

\begin{table}[H]
\centering
\begin{tabular}{|l|}
\hline
\textbf{Question (English Translation)}                    \\ \hline
Highest educational level of the mother                    \\ \hline
Highest educational level of the father                    \\ \hline
Number of refrigerators in the house                       \\ \hline
Number of computers (or laptops) in the house              \\ \hline
Number of bedrooms in the house                            \\ \hline
Number of televisions in the house                         \\ \hline
Number of bathrooms in the house                           \\ \hline
Number of cars in the house                                \\ \hline
Number of Wi-Fi networks in the house                      \\ \hline
Number of microwave ovens in the house                     \\ \hline
Number of washing machines in the house                    \\ \hline
Number of freezers in the house                            \\ \hline
Number of vacuum cleaners in the house                     \\ \hline
\end{tabular}
\label{tab:featuresNSE}
\end{table}

\section{Appendix B - Prompt for Multiple-choice Questions}
\label{appendix:multiplechoice-prompt}

\begin{table}[H]
\noindent\fbox{%
    \parbox{\columnwidth}{%
       You are designed to answer the following multiple
choice question. Without adding any extra characters,
spaces, or newline characters in the answer, provide a
single alternative as the answer”.

\begin{center}
\textbf{[QUESTION]}
\end{center}
       }}
    \caption{Prompt fed to the LLMs for each multiple-choice task (translated from Portuguese).}
    \label{tab:multiplechoice-prompt}
\end{table}

\section{Appendix C - Prompts for Essay Generation}
\label{appendix:essay-prompt}

\begin{table}[H]
\noindent\fbox{%
    \parbox{\columnwidth}{%
       \begin{center}
        \textbf{[SAMPLE TEXTS]}
       \end{center}
       
       Based on the sample texts above, write an ENEM-style essay about ``Challenges in valuing traditional communities and peoples in Brazil" in 500 words. Along the essay, make use of cultural references, such as books and philosophers. Finally, your conclusion must include a paragraph detailing agent, action, means and aim. In total, your text must have only four paragraphs.
       }}
    \caption{Prompt for essay generation (translated from Portuguese). Includes explicit system prompting and sample texts .}
    \label{tab:essay-prompt1}
\end{table}

\begin{table}[H]
\noindent\fbox{%
    \parbox{\columnwidth}{%
       Write an ENEM-style essay about ``Challenges in valuing traditional communities and peoples in Brazil" in 500 words. Along the essay, make use of cultural references, such as books and philosophers. Finally, your conclusion must include a paragraph detailing agent, action, means and aim. In total, your text must have only four paragraphs.
       }}
    \caption{Prompt for essay generation (translated from Portuguese). Includes explicit system prompting, but does not include sample texts.}
    \label{tab:essay-prompt2}
\end{table}

\begin{table}[H]
\noindent\fbox{%
    \parbox{\columnwidth}{%
       \begin{center}
        \textbf{[SAMPLE TEXTS]}
       \end{center}
       
       Based on the sample texts above, write an essay about ``Challenges in valuing traditional communities and peoples in Brazil".
       }}
    \caption{Prompt for essay generation (translated from Portuguese). Does not include explicit system prompting, but includes sample texts.}
    \label{tab:essay-prompt3}
\end{table}

\begin{table}[H]
\noindent\fbox{%
    \parbox{\columnwidth}{%
       Write an essay about ``Challenges in valuing traditional communities and peoples in Brazil".
       }}
    \caption{Prompt for essay generation (translated from Portuguese). Does not include explicit system prompting or sample texts.}
    \label{tab:essay-prompt4}
\end{table}

\section{Appendix D - Description of Socioeconomic Status (SES) Levels}
\label{appendix:level_descriptions.}

\begin{table}[H]
\centering
\small
\begin{tabular}{|c|p{6.4cm}|}
\hline
\textbf{Level} & \textbf{Description} \\ \hline
I & The majority of students have unit values of goods, and their parent’s education ranges from incomplete middle school to complete middle school. \\ \hline
II & The majority of students have unit values of goods, and their parent’s education ranges from incomplete middle school to complete high school. \\ \hline
III & The majority of students have unit values of goods, with some additional services, and their parent’s education ranges from complete middle school to complete high school. \\ \hline
IV & Most students have broader values for goods and services, and their parent’s education ranges from complete middle school to complete high school. \\ \hline
V & Most students have broader values for goods and services, and their parent’s education ranges from complete middle school to complete college degree. \\ \hline
VI & Most students have broader values for goods and services, and their parent’s education ranges from complete high school to complete college degree. \\ \hline
VII & The majority of students have broader and more diverse values of goods and services, and their parent’s education ranges from complete high school to complete college degree. \\ \hline
VIII & The majority of students have the highest values for goods and services, and their parent’s education is dominated by complete college degrees. \\ \hline
\end{tabular}
\caption{Socioeconomic Level Descriptions}
\label{tab:level_descriptions.}
\end{table}

\section{Appendix E - Factors that Affect Student Performance}

The decision to analyze the socioeconomic level of students was based on Brazilian research involving the National High School Exam (ENEM) which reveals that the factor that most affects the students' average is precisely income~\cite{travitzki2014desigualdades}, as previous research has noted, every 5\% increase in the number of students from high-income families leads to an increase of about 6.25 points in the multiple-choice score in ENEM~\cite{renda_enem}. Following this reasoning it was decided that the most important factor to be taken into account in the present study is the student's socioeconomic level, rather than gender or ethnicity.

\section{Appendix F - The Divergence Between the Multiple Choice Questions and the Essay}
\label{appendix:multiple_choice_vs_essay}

Since divulging the actual essays of students may hurt the anonymity of the data, INEP only provides the answers of each candidate for the multiple choice questions and the final grades. Thus, we only have the texts of students who chose to publicize their essays, limiting the sample size.

\end{document}